\documentclass[letterpaper, 10 pt, conference]{ieeeconf}

\IEEEoverridecommandlockouts                              

\usepackage{amsmath} 

\usepackage{algorithm,algpseudocode}
\usepackage{amsfonts}

\usepackage{amssymb}
\newcommand{\R}{\mathbb{R}}

\title{\LARGE \bf Robotic Laser Orientation Planning with a 3D Data-driven Method}

\author{Guangshen Ma $^{1}$, Weston Ross $^{2}$ and Patrick J. Codd $^{1, 2}$
\thanks{$^{1}$ Department of Mechanical Engineering and Materials Science, Duke University.}%
\thanks{$^{2}$ Department of Neurosurgery, Duke University Medical Center.}
\vspace{-3mm}
}

\usepackage{lipsum}
\usepackage{graphicx}

\usepackage{lipsum}
\usepackage{adjustbox}
\usepackage{comment}
\usepackage{xcolor}
\usepackage{float}
\usepackage{caption}
\usepackage{subcaption}
\usepackage{tabularx}
\usepackage[english]{babel}
\usepackage[utf8]{inputenc}
\usepackage{algorithm}
\usepackage{amsmath}
\usepackage{hhline}

\DeclareMathOperator*{\argmin}{arg\,min}

\begin{document}

\maketitle
\thispagestyle{empty}
\pagestyle{empty}

\begin{abstract}

This paper focuses on a research problem of robotic controlled laser orientation to minimize errant over-cutting of healthy tissue during the course of pathological tissue resection. Laser scalpels have been widely used in surgery to remove pathological tissue targets such as tumors or other lesions. However, different laser orientations can create various tissue ablation cavities, and incorrect incident angles can cause over-irradiation of healthy tissue that should not be ablated. This work aims to formulate an optimization problem to find the optimal laser orientation in order to minimize the possibility of excessive laser-induced tissue ablation. We first develop a 3D data-driven geometric model to predict the shape of the tissue cavity after a single laser ablation. Modelling the ``target" and ``non-target" tissue region by an obstacle boundary, the determination of an optimal orientation is converted to a collision-minimization problem. The goal of this optimization formulation is maintaining the ablated contour distance from the obstacle boundary, which is solved by Projected gradient descent. Simulation experiments were conducted and the results validated the proposed method with conditions of various obstacle shapes and different initial incident angles. 

\end{abstract}

\section{INTRODUCTION AND RELATED WORKS}

Robotic laser scalpels have been widely used in different surgical tasks, such as eye surgery \cite{channa2017robotic, MonocularLaser}, neurosurgery \cite{liao2012integrated, ross2018automating} and dermatology \cite{penza2019hybrid}. Precise control of laser energy delivery to tissues ensures optimal treatment of targeted lesions and tissue regions, making robotic control of these systems an essential element for precision laser guided surgery.  A single laser pulse can create a 3D volumetric cavity on the tissue surface, and its shape is related to the laser incident angle and laser energy delivery.  Incorrect orientation planning and laser energy delivery can result in damage to collateral "non-target" (healthy) tissue removal.  Therefore, a natural question arises to whether there exists an optimal laser orientation to minimize the probability of incorrect laser energy deliver. Calculating the optimal orientation angle for a single laser pulse thus becomes a vitally important problem in robotic laser surgical planning.


Robotic laser orientation problems have been widely studied in robotic laser cutting \cite{dolgui2009manipulator}, industrial robotic manipulation \cite{fang2013orientation} and robotic laser surgery \cite{penza2019hybrid, burgner2010ex}, but not for the proposed application of minimizing errant tissue overcutting. These methods generally leverage the vision or user inputs to develop an orientation planning strategy. The laser scalpel has been typically affixed to the 6-DOF (degree-of-freedom) robotic arm end-effector, with the robot controlling the laser in order to move towards a predefined planning trajectory. Such robotic orientation planning can improve the safety and robustness of these surgical systems \cite{dolgui2009manipulator, fang2013orientation, jivraj2019robotic}. For example


\begin{figure}[h]
\centering
\includegraphics[scale = 0.42]{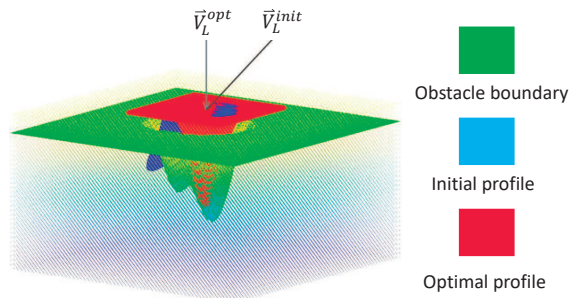}
\caption{3D ablated contours created by two laser orientations.}
\label{fig1_3ddemo}
\end{figure}


\noindent
in Fig.~\ref{fig1_3ddemo}, different 6-DOF laser orientations can create distinct cavities and causes various effects for laser surgery. 


Therefore, we proposed a research question as whether it exists an optimal laser incident orientation to minimize the distance between the ablated profile and a pre-defined obstacle boundary? This boundary can be defined as either a preoperative ``edge" between healthy and non-healthy tissue, or a pre-designed cavity contour with a fixed geometry. This research includes two major sub-problems:

\subsection{Creating the Laser-tissue Geometric Model}

The laser-tissue geometric model refers to a kinematic system capable of predicting the ablated tissue contour under a given laser incident angle. Given a 3D point on the tissue surface, this model predicts the depth-of-cut, and thus estimates the new position after tissue removal. The major modelling approaches can be summarized as Model-based and Model-free methods. 

Model-based method refers to estimating the ablated contour by calculating the energy delivery for each tissue position. Stopp \textit{et al.} proposed a laser-tissue ablation model to create pre-defined 3D geometries for robotic bone surgery \cite{stopp2008new}. This method builds a mathematical model to estimate the energy delivered to a 3D tissue position (modelled as 3D voxel in simulation) and predicts the resulting ablated contour. However, modelling the complex physics for laser-tissue interaction (thermal effect, optical property) is difficult since it is prone to different experimental conditions such as water spray, surface geometry and tissue material.

Model-free method aims to model the tissue of removal by using an entirely data-driven approach. As it is difficult to model the laser-tissue interaction due to the heterogeneity of tissue material and the complex physical mechanism, prior works have examined applying a data-driven method to model the physics \cite{fichera2016online, burgner2010ex, kahrs2010planning, kahrs2008visual, stopp2008new}. The laser beam profile can be usually modelled by a Gaussian function \cite{walsh1988pulsed, ross2018optimized}. The tissue of removal should follow the similar pattern since the depth-of-cut is related to the strength of energy delivered to the target. Therefore, the Gaussian-based model has been widely used to describe the laser-tissue relation, and the parameters of the Gaussian function can be learned through the 3D cavity data collected by high-resolution scanners such as confocal microscopy \cite{kahrs2010planning, kahrs2008visual} and computed tomography (CT) \cite{ross2018optimized}.

The method of learning laser-tissue physics by Gaussian function fitting has also been applied in various robotic laser applications, such as tissue depth control \cite{fichera2016online}, surgical simulation \cite{burgner2010ex, kahrs2010planning} and generating a cutting path for volumetric resection \cite{ross2018optimized}. However, these studies have not discussed the problem of orientation planning with various laser angles and the application controlling the ablated profiles for robotic laser surgery, which inspires the proposed work.


\subsection{Modeling the Optimal Laser Orientation Problem}

\begin{figure}[h]
\centering
\includegraphics[scale = 0.35]{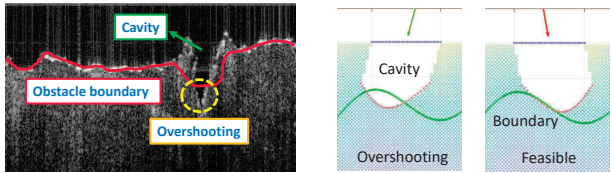}
\caption{2D laser-tissue model. (a) shows an ablated contour outside the labelled boundary (overshooting). (b) denotes the two types of tissue removal.}
\label{fig2_2dexp}
\end{figure}

The optimal laser orientation problem can be modelled as an ``Obstacle avoidance" planning problem by which the ablated contour, as created by the current laser angle, should keep the adequate distance to the pre-defined boundary (between ``target" and ``non-target" tissue region). For example in Fig.~\ref{fig2_2dexp} (a), an ablated contour of a single laser pulse is located outside of a pre-designed boundary, which can be measured in OCT slice image. These images can be concatenated to formulate a 3D point cloud \cite{draelos2021contactless, draelos2018real} and thus the 2D obstacle boundary can become a 3D surface.

Given a pre-defined obstacle boundary and a laser-tissue geometric model, the goal is to find an optimal laser orientation that creates a collision-free ablation contour. Fig.~\ref{fig2_2dexp} (b) depicts two examples of over-cutting and feasible ablation in 2D. Modelling the 3D obstacle avoidance problem is an important topic in optimization-based motion planning \cite{zucker2013chomp, schulman2014motion} and these well-studied methods can be applied in this work for collision modelling. The obstacles are usually modelled by the Euclidean distance transforms (EDT) and the signed distance \cite{zucker2013chomp, schulman2014motion}, which can be used to formulate a 3D voxel field that encodes the collision information (distance to the obstacle boundary). These distance fields can be employed to guide the orientation planning.

This research is inspired by the ideas of laser-tissue modeling in surgical planning and collision avoidance in robotics. We propose a data-driven geometric model to describe the laser-tissue cavity and a novel optimization problem is formulated to find the optimal laser orientation. This method opens novel potential applications for conducting surgical simulation with different 3D sensor guided robotic laser platforms, such as stereovision \cite{schoob2015tissue}, RGB-D camera \cite{ma2019novel} and optical coherence tomography (OCT) \cite{draelos2021contactless}. 






\section{METHODS}

\subsection{Laser-tissue Geometric Model}

\subsubsection{Gaussian Profile}

\begin{figure}[h]
\centering
\includegraphics[scale = 0.36]{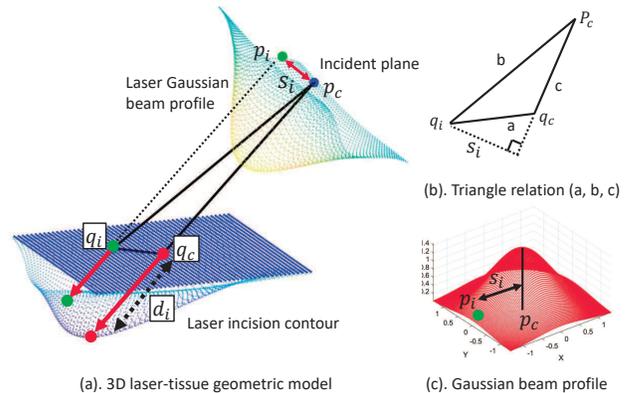}
\caption{3D laser-tissue geometric model, laser-surface geometry and the Gaussian beam profile of $CO_2$ laser scalpel.}
\label{3dgpgeo}
\end{figure}

This work proposed a laser-tissue model of the $CO_2$ laser with a wavelength of $10.6~\mu m$ and a $1/e^2$ spot size of 0.80 mm, which our group has utilized in several studies of laser ablation \cite{ross2018automating, ma2019novel, tucker2021creation}. The $CO_2$ beam profile can be generally described by a Gaussian function \cite{leung2012real, acemoglu2017laser, bay2015real}. Fig.~\ref{3dgpgeo} (a) illustrates the complete 3D geometric model. The depth-of-cut for a surface position depends on the strength of the laser energy and thus follows the Gaussian pattern. Fig.~\ref{3dgpgeo} (b) and (c) illustrate the geometric configuration of the laser-surface model and the Gaussian-shape $CO_2$ beam profile. While a $CO_2$ laser is utilized for demonstrative purposes, this approach can be generalized to other laser scalpels that likewise have Gaussian beam profiles.

To model the laser-tissue geometry, we first define a laser ``incident plane" described by the laser center and the orientation vector, which has the same shape as the Gaussian beam profile. The surface points can be projected to this plane to estimate the depth-of-cut. A region of interest (ROI) is defined on the superficial tissue surface. In this configuration, $\mathbf{q_c} \in \mathbb{R}^3$ denotes the laser ablation center on the 3D surface and $\mathbf{q_i} \in \mathbb{R}^3$ is a query point around this center. We have the relation: 

\begin{equation}
    \label{eq_1}
    \begin{aligned}
        \mathbf{p_c} & = \mathbf{q_c} - \mathbf{v} * L_{ref}
    \end{aligned}
\end{equation}

Where $\mathbf{p_c} \in \mathbb{R}^3$ is defined as the laser incident center and $\mathbf{v} \in \R^3$ is the incident vector. The operator $||\cdot||_2$ refers to the L2-norm and we usually restrict $||\mathbf{v}||_2 = 1$. $L_{ref}$ is a reference distance that can be set as an arbitrary constant, since the geometric configuration does not depend on this value (set as ``1" ).

The depth-of-cut $d_i$ for the surface point $\mathbf{q_i}$ is defined as the tissue removal at the incident direction. Based on the Gaussian beam profile assumption, $d_i$ can be calculated by the projected distance $s_i$ between $\mathbf{p_i}$ and $\mathbf{p_c}$: 

\begin{equation}
    \label{eq_2}
    \begin{aligned}
      d_i & = L_{G} * \mathrm{exp}( \frac{s_i^2( \mathbf{v}, \mathbf{q_i})}{-2 * \sigma_{G}^2})
    \end{aligned}
\end{equation}

Where $\mathrm{exp}(\cdot)$ is an exponential operator. We assume the Gaussian beam profile is symmetric \cite{kahrs2008visual} and the parameters of $L_{G}$ and $\sigma_{G}$ can be estimated by the data-driven method for a specified tissue material and laser setting. Another important task is to determine $s_i$. Fig.~\ref{3dgpgeo} (b) explains the triangle configuration among $\mathbf{p_c}$, $\mathbf{q_i}$ and $\mathbf{q_c}$. The $s_i$ is the altitude of the triangle with the sides of $\mathbf{p_c}$ and $\mathbf{q_c}$: 

\begin{equation}
    \label{eq_3}
    \begin{aligned}
        \frac{s_i * c}{2} & = \sqrt{p(p-a)(p-b)(p-c)} \\
    \end{aligned}
\end{equation}

Where $\sqrt{p(p-a)(p-b)(p-c)}$ is the triangle area denoted by the three sides $a, b, c$, and we have $p = \frac{a + b + c}{2}$. 
 
\subsection{Data-driven Method for Gaussian Function Parameters}

Given a laser-tissue geometric model, we need to estimate the amplitude $L_{G}$ and variance parameter $\sigma_{G}$. In practice, the amplitude is a function of laser power, time, and the optical properties of the tissue that drive ablation.  Herein, we assume that any ablation amplitude is achievable by tuning the laser parameters and thus we can consider it as a controllable variable. The $L_{G}$ and $\sigma_{G}$ can usually be function-fitted with the 3D data collected by high-resolution scanner \cite{burgner2010ex, fichera2016online, kahrs2010planning}. Our prior work in \cite{ma2020characterization} has validated the feasibility of using Micro-CT data to characterize the laser-tissue cavities under various incident angles but was limited to 2D analysis. This study extends the analysis from 2D to 3D, building on the same Micro-CT data to determine the Gaussian function parameters. 



\begin{figure}[h]
\centering
\includegraphics[scale = 0.32]{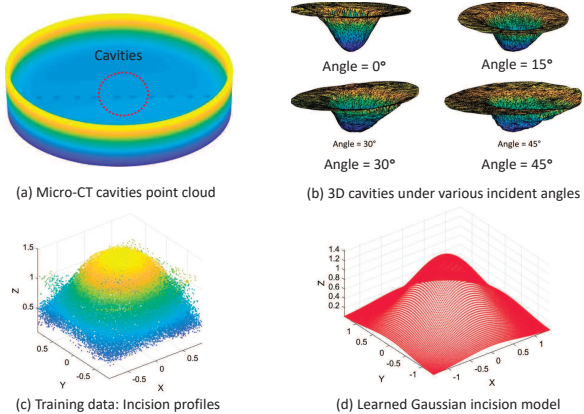}
\caption{Parameters estimation by Micro-CT data.}
\label{3ddatadriven}
\end{figure}


We have a dataset of 3D measurements from a list of 3-DOF orientation angles $\{ \boldsymbol{ \theta_i } \} \in \mathbb{R}^3 $. Each laser pulse at $ \theta_i $ can create a tissue cavity described by the Micro-CT point cloud data \cite{ma2020characterization}, which is defined as $\{ \boldsymbol {q_{i,j}} \} \in \mathbb{R}^3$. The \textit{i} refers to the index of incident angle and \textit{j} as the index of measurement. The point cloud can be converted to the projected coordinates at the laser incident plane and this is referred as $\{ \boldsymbol{ s_{i,j} } \} \in \mathbb{R}^2 $. With each $s_{i,j}$, we can calculate the depth-of-cut $d_{i,j}$ by Equation.~\ref{eq_2} and obtain $\{ \boldsymbol {d_{i,j}} \} \in \mathbb{R}^1 $. Therefore, we obtain a dataset $\{ \boldsymbol{ s_{i,j}, d_{i,j} } \} \in \mathbb{R}^3$, which can be used to estimate the Gaussian function parameters $(L_{G}, \sigma_{G})$. The function-fitting is achieved by using the log operation and the non-linear least-squares fitting metric \cite{guo2011simple}. Fig.~\ref{3ddatadriven} illustrates the Gaussian parameters fitting procedure.

\subsection{Laser-tissue Kinematic System}

The laser-tissue geometric model can be referred as a kinematic system that describes the relation between the laser incident angle and the resulting ablation contour. For a given incident direction, this contour is formulated by the surface sampled points after the tissue removal. For each surface point $\mathbf{q}_{i}^{k} \in \mathbb{R}^3$ at k-th time step, we have:

\begin{equation}
    \label{eq_4}
    \begin{aligned}
        Q(\mathbf{v}, \mathbf{q}_{i}^k) =   \mathbf{q}_{i}^{k+1}(\mathbf{v}, \mathbf{q}_{i}^k) & = \mathbf{q}_{i}^{k} + \mathbf{v} * d_i
    \end{aligned}
\end{equation}

Where we have $d_i = L_G * \mathrm{exp} ( \frac{s_i^2(\mathbf{v}, \mathbf{q}_i^k)}{- 2 * \sigma_{G}^2} )$ from Equation.~\ref{eq_2}. $\mathbf{q}_{i}^{k+1} \in \mathbb{R}^3$ is the updated position from $\mathbf{q}_{i}^{k}$. The $L_{G}$ and $\sigma_{G}$ are the learned Gaussian function parameters. The $s_i(\mathbf{v}, \mathbf{q}_i^k)$ depends only on $\mathbf{v}$ and $\mathbf{q}_i$. This kinematic model can be used to predict a new surface profiles given the initial surface point cloud and the incident angle.


\subsection{Collision Modelling with Obstacle Boundary}

\subsubsection{EDT Field} In this study, the main goal is to find an optimal laser orientation that can maintain the distance between the ablated contour and the obstacle boundary. We model the obstacle boundary by the Euclidean distance transforms (EDT), as has been widely used for optimization-based motion planning of creating a collision-free path \cite{zucker2013chomp}. EDT represents the distance to the nearest obstacle position for each 3D voxel \cite{felzenszwalb2012distance}, which can be used to measures the level of collision for an arbitrary 3D position \cite{felzenszwalb2012distance}.


\subsubsection{Obstacle Modelling} 
\begin{figure}[h]
\centering
\includegraphics[scale = 0.40]{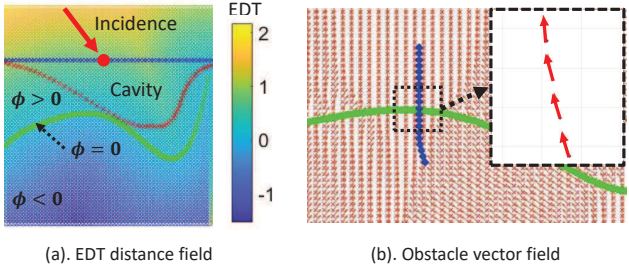}
\caption{2D EDT value distribution and obstacle vector field.}
\label{2dedt}
\end{figure}

The obstacle boundary can be considered as a 3D surface where the EDT value is zero \cite{zucker2013chomp, schulman2014motion}. Specifically, we define $\Phi(\cdot)$ as the EDT levelset function and $\Phi(\cdot) = 0$ denotes the boundary positions. $\Phi(\cdot) > 0$ denotes the target region and  $\Phi(\cdot) < 0 $ denotes the non-target region, as shown in Fig.~\ref{2dedt} (a). A EDT vector field $\nabla \Phi(\cdot)$ can be calculated by the directional gradients with the 3D Sobel gradient operator \cite{gonzalez2002digital}. $\nabla \Phi(\cdot)$ denotes a direction of the lower collision cost and thus can be used to guide the orientation planning. Similar to \cite{zucker2013chomp}, the obstacle cost $C(\Phi(x))$ is defined as a piecewise function: 

\begin{equation}
    \label{eq_5} C(\Phi(x)) = 
    \begin{cases} 
      -\Phi(x) + \frac{1}{2} \epsilon & x \leq 0 \\
      \frac{1}{2\epsilon} (\Phi(x) - \epsilon)  & 0 < x \leq \epsilon \\
      0 & x > \epsilon
   \end{cases}
\end{equation}

Where $\epsilon$ is the obstacle distance threshold. $\Phi$ is the EDT value and $x$ is the 3D voxel position. The gradient of the cost function $\nabla C (\Phi) = \frac{\partial C}{\partial \Phi}$ is negative if the position is getting closer to the obstacle boundary, which will guide the point to move to a reverse direction. Fig.~\ref{2dedt} (a) illustrates the EDT values distribution in 2D. The EDT can be set as positive and negative inside and outside the obstacle boundaries. Fig.~\ref{2dedt} (b) depicts the vector trajectory around the boundary, which can guide the point movement towards the positive EDT region.  


\vspace{-3mm}

\subsection{Orientation Planning Algorithm}

The goal of the orientation planning is to minimize the total obstacle costs for the ablated contour. The contour can be represented by a point cloud data that can be collected by any intraoperative 3D sensors. In practice, a laser incident orientation is usually perpendicular to the point of interest, and is only allowed to adjust to a small angle. We can formulate a constrained optimization problem:

\begin{equation}
    \begin{aligned}
      \min_{ \boldsymbol{ \theta } \in \mathbb{R}^3} \quad f = & \sum_i C ( \Phi ( Q( \mathbf{v} (\boldsymbol{\theta}), \mathbf{q}_{i} ) ) )  \\
      \textrm{s.t.} \quad & \boldsymbol{\theta_1} \leq \boldsymbol{\theta} \leq \boldsymbol{\theta_2}
    \end{aligned}
    \label{eq_6}
\end{equation}


This objective function $f(\cdot)$ denotes the summation of obstacle costs for each point at the ablated contour. We also have the incident orientation $\mathbf{v}(\boldsymbol{\theta})$ controlled by the angle vector $\boldsymbol{\theta} \in \mathbb{R}^3$. The $C(\cdot)$ represents the obstacle cost at the current 3D voxel position. $\Phi(\cdot)$ is the EDT distance function. $Q(\cdot)$ is the function operator following the kinematic model in Equation.~\ref{eq_4}. $\boldsymbol{\theta_1}$ and $\boldsymbol{\theta_2}$ describe the 3-DOF angle range. As the ablation center $q_c$ is fixed, the laser orientation $\mathbf{v}(\boldsymbol{\theta}) $ is related to the incident angle and the surface point $\mathbf{q}_i$. 

\subsubsection{Analytical gradient}

The analytical gradient of the objective function is important for the optimization solver and can be derived by chain rule. $\nabla C(\cdot)$ and $\nabla \Phi(\cdot)$ can be obtained by Equation.~\ref{eq_5} and the EDT field. For $Q(\cdot)$, we first simplify and rewrite the kinematic system model:


\begin{equation}
    \begin{aligned}
        Q(\cdot) & = \mathbf{q}_{i} + \mathbf{v} (\boldsymbol{\theta}) * L_{G} * \mathrm{exp} ( \mu( \mathbf{v} (\boldsymbol{\theta}) ) )
    \end{aligned}
    \label{eq_7}
\end{equation}

Where $\mu(\mathbf{v} (\boldsymbol{\theta})) =  \frac{ s_i^2( \mathbf{v} (\boldsymbol{\theta}, \mathbf{q}_i) ) }{ -2 * \sigma_{G}^2 }$. To take the derivative of $Q(\cdot)$ with respect to the incident angle $\boldsymbol{\theta}$: 

\begin{equation}
    \begin{aligned}
       \quad \mathbf{J} & = \frac{\partial Q}{\partial \mathbf{\theta}} = \frac{\partial Q}{\partial v } * \frac{\partial v } {\partial \mathbf{ \theta  } } \\ 
        \quad & = 
        \begin{bmatrix}
        \frac{\partial Q_1}{\partial \theta_1} & \frac{\partial Q_1}{\partial \theta_2} & \frac{\partial Q_1}{\partial \theta_3} \\
        \frac{\partial Q_2}{\partial \theta_1} & \frac{\partial Q_2}{\partial \theta_2} & \frac{\partial Q_2}{\partial \theta_3} \\
        \frac{\partial Q_3}{\partial \theta_1} & \frac{\partial Q_3}{\partial \theta_2} & \frac{\partial Q_3}{\partial \theta_3} \\
        \end{bmatrix}
    \end{aligned}
    \label{eq_8}
\end{equation}

The $\mathbf{J}$ is the Jacobian matrix of the system model and each item is $\frac{\partial Q_i}{\partial v_j} = \sum_{t = 1,2,3} \frac{\partial Q_i}{\partial v_t} * \frac{\partial v_t}{\partial \theta_j}$ (\textit{i} and \textit{j} are index of item in $\mathbf{J}$). We also have (complete details in \textbf{Appendix}): 


\begin{equation}
    \begin{aligned}
        \frac{\partial Q}{\partial v_j}  & = \frac{\partial q_i}{\partial v_j} + \frac{v_i}{v_j} * L_{G} *  exp( \mu( \cdot ) ) \\
        & + \frac{\partial }{\partial v_j} ( exp( \mu( \cdot ) ) * L_{G} * v_i ) \\
        \frac{\partial v_j}{\partial \theta_j } & = \frac{\partial }{\partial \theta_j} ( R_x(\theta_x) * R_y(\theta_y) * R_z(\theta_z) ) * \mathbf{v}_0  
    \end{aligned}
    \label{eq_8}
\end{equation}

Where $R_x$, $R_y$ and $R_z$ are the rotation angles that control the initial orientation vector $\mathbf{v}_0$. The $\frac{\partial v_j}{\partial \theta_j }$ can be easily obtained by taking the derivative for the rotation matrix. The $ \mathrm{exp} ( \mu( \cdot ) ) = \mathrm{exp}( \frac{ s_i^2( \mathbf{v} (\boldsymbol{\theta}, \mathbf{q}_i) ) }{ -2 * \sigma_{G}^2 } ) $ is a scalar function of $s_i(\cdot)$. The gradient of $s_i(\cdot)$ is related to the three sides $a(\cdot) = ||q_c - q_i||_2$, $b(\cdot) = q_i - q_c + \mathbf{v(\theta)} * L_{ref}$ and $c(\cdot) = p_c - q_c = L_{ref}$, as depicted in the triangle configuration (Fig.~\ref{3dgpgeo}). Thus, the gradient can be derived by chain rule (complete details in \textbf{Appendix}). Finally, we obtain the analytical gradient $\nabla f(\boldsymbol{\theta}, \mathbf{q}_i)$ for the objective function: 

\begin{equation}
    \begin{aligned}
      \nabla f(\boldsymbol{\theta}, \mathbf{q}_i) & = \sum_i \frac{\partial C}{\partial \Phi} * 
      \frac{\partial \Phi}{\partial Q} * \frac{\partial Q (\boldsymbol{\theta}, \mathbf{q}_i)}{\partial \mathbf{\theta}} 
    \end{aligned}
\end{equation}

Where $\frac{\partial Q (\boldsymbol{\theta}, \mathbf{q}_i)}{\partial \mathbf{\theta}}$ is the Jacobian matrix in Equation.~\ref{eq_8} and can be calculated from Equation.~\ref{eq_8}.

\subsubsection{Projected Gradient Descent}


For the incident angle constraint, we use the Projected gradient descent method (Proj-GD) to solve the optimization problem \cite{nocedal2006numerical}. For each iteration, Proj-GD first computes the gradient update:

\begin{equation}
    \begin{aligned}
      \boldsymbol{ \theta^{k + 1} } & = \boldsymbol{ \theta^k } - \alpha * \nabla f ( \boldsymbol{\theta} ) 
    \end{aligned}
    \label{eq_9}
\end{equation} 

Where $\alpha$ is the step size for the gradient update. For the angle constraint, Proj-GD finds a new update by solving an quadratic optimization problem \cite{nocedal2006numerical}:

\begin{equation}
    \begin{aligned}
      \boldsymbol{\theta_*}^{k+1} = \mathbf{Proj} (\boldsymbol{\theta}) & = \argmin_{\boldsymbol{\theta_1} \leq \boldsymbol{\theta} \leq \boldsymbol{\theta_2}} \frac{1}{2} ||\boldsymbol{\theta} - \boldsymbol{\theta^{k+1}}||_2
    \end{aligned}
    \label{eq_10}
\end{equation}

Where $\boldsymbol{\theta}^{k+1}$ is from Equation.~\ref{eq_9} and $\boldsymbol{\theta_*}^{k+1}$ denotes the final gradient after Proj-GD. The $\mathbf{Proj}(\cdot)$ is the operator for Projected gradient descent and it represents the nearest feasible point to gradient-updated position. An example of Proj-GD is depicted in Fig.~\ref{explong} (a).

\subsubsection{Data selection for each iteration}

The $\nabla f(\boldsymbol{\theta})$ denotes the summation of gradients for each data point. Each gradient depends on the unique configuration of the surface position $q_i$ and the incident direction, which can be calculated differently. It is likely that some gradients are contradicted and pointing at the reverse directions and a 2D reverse gradient example is depicted in Fig.~\ref{exp12d-A} (b). This will cause inaccurate updates and the ablated contour would be stuck in a local and sub-optimal region. Another problem is that some sample points will move to the positions with zero costs and we need to select these points during the optimization. Therefore, we propose a data selection approach to solve both problems. For each point robot, we calculate the reduction gain of the collision cost $C_i$ based on the updated angle $\boldsymbol{\theta}_*^{k+1}$: 

\begin{equation}
    \begin{aligned}
      gain(\boldsymbol{\theta^k}) & = C( \boldsymbol{\theta_*^{k+1}} ) - C( \boldsymbol{\theta^{k} }  )
    \end{aligned}
    \label{eq_11}
\end{equation}

For each iteration, we rank the gain with an increasing order and select the Top-N sample data (N can be set as 30\% of the sample data). This method ensures the adequate cost reduction and guarantees that effective sample points could be selected to guide the laser orientation planning.

\section{Simulation Experiments}



\subsection{Data-driven Model with Micro-CT Data}


We first validate the data-driven method by using the Micro-CT data, as discussed in our prior work \cite{ma2020characterization}. We collected the 3D cavity point cloud data from 4 incident angles with 10 repeated measurements. This formulates a dataset for the estimation of Gaussian function parameters $L_{G}$ and $\sigma_G$. The Gaussian function fitting is a non-linear curve fitting problem (least-squares minimization) and we use the MATLAB ``lsqcurvefit" optimization toolbox to find the fitted parameters \cite{coleman1996interior}. For the specified tissue material and laser parameters, we obtain $L_{G} = 1.4376$ and $\sigma_G = 0.6486$. The model re-projected Root-mean-square error is $0.1468$ and this contributes to $10.2\%$ of the $L_{G}$. This shows the laser can generate a maximum incision depth with around 1.44 mm at the ablation center. For generality of the problem, we can build a loop-up tables of laser-tissue models and parameters which can be used directly in Robotic surgery.


\subsection{Test 1: Validate the Analytical Gradient}



\textit{Test 1} aims to validate the analytical Jacobian matrix $\nabla Q(\boldsymbol{\theta})$ by making a hypothesis that the update of the orientation can follow the EDT gradient vectors $\nabla \Phi(Q(\boldsymbol{\theta}))$. The $\nabla \Phi(Q(\boldsymbol{\theta}))$ refers to a vector direction with higher EDT values with respect to a surface position $Q(\boldsymbol{\theta})$, which indicates a lower collision cost. For example, if we have a vector $\nabla \Phi(Q(\boldsymbol{\theta})) > 0$ and use the gradient ascent $\boldsymbol{\theta^{k+1}} = \boldsymbol{\boldsymbol{\theta}^{k}} + \alpha * \nabla_{\theta} \Phi (Q(\theta^k))$ as the update rule, we obtain $\Phi^{k+1} = \Phi^k + \frac{ (\boldsymbol{\theta}^{k+1} - \boldsymbol{\theta}^{k})^2 } { \alpha } + \Phi^k \geq \Phi^k$. This indicates that if the angle is updated by the gradient ascent direction, the EDT value can increase and the ablated contour can maximize the distance to the obstacle boundary (minimize cost). We can thus conclude that the gradient ascent with $\nabla \Phi(\boldsymbol{\theta})$ is equal to the gradient descent with $\nabla C(\boldsymbol{\theta})$, because $\nabla C(\boldsymbol{\theta})$ is negative when the EDT value is smaller than the collision threshold. For brevity, we use the $\nabla \Phi(\cdot)$ with the gradient ascent rule to validate the feasibility of the analytical gradient. 

For the simulated experiments, we define a ``point robot" for a surface position $\mathbf{q}_i$ which follows the kinematic model in Equation.~\ref{eq_4}. The ``point robot" refers to the change of movement of a surface position after a single laser incidence. The validation of the analytical Jacobian matrix $\nabla Q(\boldsymbol{\theta})$ can be achieved by observing the movement of the point robot. 


\begin{figure}[h]
\centering
\includegraphics[scale = 0.40]{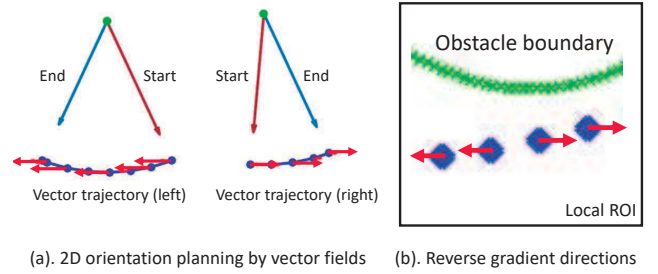}
\caption{2D Orientation planning by reference vector.}
\label{exp12d-A}
\end{figure}

\begin{figure}[H]
\centering
\includegraphics[scale = 0.36]{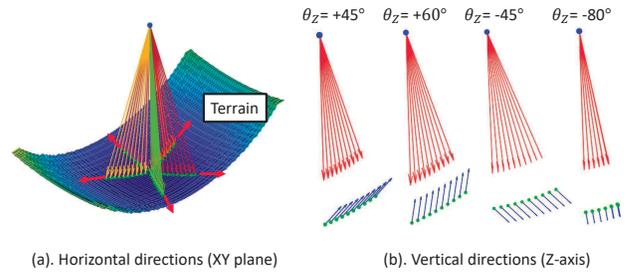}
\caption{Test 1: Orientation planning in 3D.}
\label{exp13dxyz}
\end{figure}

We first conduct a 2D test to show that the 1-DOF orientation angle update can be guided by a reference direction. Fig.~\ref{exp12d-A} (a) illustrates that the incident orientation is guided by a fixed reference direction (left or right). For the 3D analysis, the orientation planning follows the gradient ascent update rule and is guided by the arbitrary 3-DOF reference vectors defined as $\mathbf{v_R} = \mathbf{v_H} + \mathbf{v_G}$. This vector is generated by setting the rotation angles in $[-180^{\circ}, 180^{\circ}]$ (horizontal) and $[-80^{\circ}, 80^{\circ}]$ (vertical). It is noted that one of the rotation angle is repeated and thus not used to create the sample vectors. For each reference vector, an incident orientation controls the point robot to generate a trajectory. Fig.~\ref{exp13dxyz} illustrates the examples of controlling the angles to follow the reference trajectories. 

To validate the method, the orientation error is used to measure the difference between the reference vector and the point vector defined as $\mathbf{v_k} = \mathbf{q}_{k+1} - \mathbf{q}_k$. As the point trajectory is located at a fixed surface terrain, as shown in Fig.~\ref{exp13dxyz} (a), we only need to consider the vector offset in XY-plane (horizontal direction). This is because the point robot cannot be controlled to a new position guided by the vertical direction. The orientation error distribution is depicted in Fig.~\ref{exp13derror} (a). These results indicate that given a reference vector fields in 3D, the point robot can be guided to follow these directions. The analytical Jacobian matrix of $\nabla Q(\boldsymbol{\theta})$ can correctly be applied to control the orientation angle guided by the EDT field.


\begin{figure*}[h]
\centering
\includegraphics[scale = 0.58]{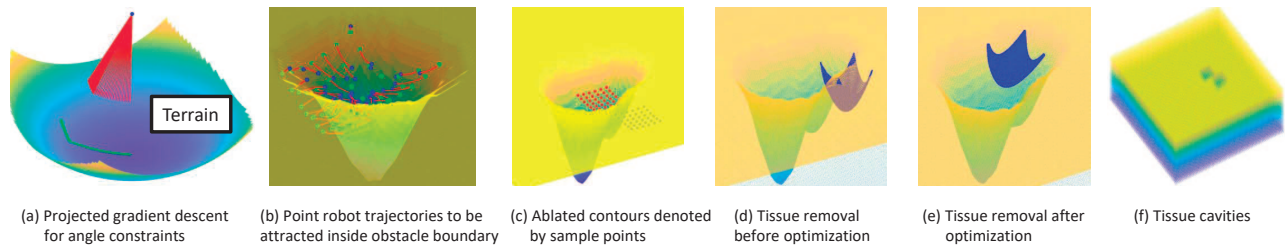}
\caption{3D diagrams of \textit{Test 2} and \textit{Test 3}. (a) shows an example case of controlling the incident angle within a range by Proj-GD. (b) shows the point robot trajectories for \textit{Test 2}. (c) to (f) show the volumetric tissue removal before and after orientation planning, with a Gaussian-shape obstacle boundary.}
\label{explong}
\end{figure*}

\subsection{Test 2: Sample Points Validation}

Instead of using a fixed reference vector, this experiment aims to validate the hypothesis that the point robots can follow a vector field created by an arbitrary obstacle boundary. It is noted that the gradient descent rule with an analytical gradient $\nabla f(\boldsymbol{\theta}, q_i)$ is used in this study. The laser-tissue model parameters are estimated by the Micro-CT data, as discussed in section III.A. 

We define 6 Gaussian-shape obstacle boundaries by setting various variance parameters and formulate the obstacle vector fields in simulation. For each obstacle terrain, we define a surface grid with different initial positions by setting 9 random incident angles. Each surface grid contains 49 point robots and this ensures that each sample point is located at an arbitrary starting position. The goal is to control each point robot to move to a low-cost position and keep adequate distance to the obstacle boundary. Fig.~\ref{explong} (b) shows an example of the point robot trajectories and all the points can successfully be ``attracted" to the inner obstacle boundary. 

\begin{figure}[h]
\centering
\includegraphics[scale = 0.26]{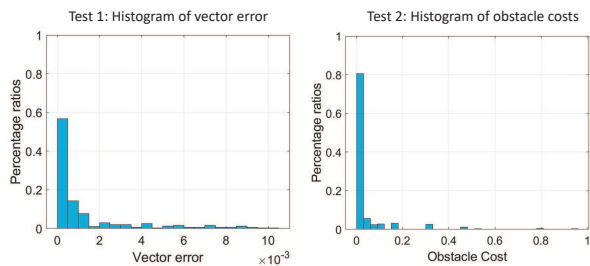}
\caption{Error histogram of $Test~1$ and $Test~2$.}
\label{exp13derror}
\end{figure}

Fig.~\ref{exp13derror} (b) illustrates the error histogram of the final obstacle costs at the last step (after optimization). The majority of the point robots can be positioned with low obstacle costs ($\leq 0.5$) based on a collision threshold as 0.6. This demonstrates the feasibility of calculating the optimal orientation angle for a single point robot movement with arbitrary initial conditions and boundary terrains.

\subsection{Test 3: Surface Validation}

\textit{Test 3} aims to validate the orientation planning method by using the gradient information from all the point robots at the surface, instead of an individual one. The experimental conditions in \textit{Test~2} are applied to this study. The Projected gradient descent and the Data selection methods are tested with the analytical gradient $\nabla f(\boldsymbol{\theta}. q_i)$. 

\begin{figure}[h]
\centering
\includegraphics[scale = 0.40]{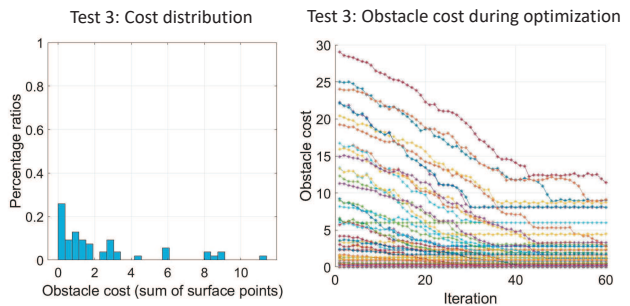}
\caption{Result analysis for \textit{Test~3}.}
\label{exp3err}
\end{figure}

The obstacle costs of all the point robots are collected for evaluation. Fig.~\ref{exp3err} (a) illustrates the cost distribution of the last step (after optimization). It shows that a high ratio of the sample surfaces can be controlled to the low-cost positions. Fig.~\ref{exp3err} (b) represents the trend of the obstacle costs for all the experimental conditions. Most cases can show the reduction of obstacle cost after the planning. For some surfaces initially localized at the correct positions, the obstacle cost has already remained at a low-cost level. It is noted that for cases where the contour of the Gaussian-shape boundary is smaller than the sample surface, the final obstacle cost cannot reach to zero or a very small value. 

Fig.~\ref{explong} (c) to (f) depict the 3D cavities before and after the optimization. The ablated contours can successfully moved to a ``safe" position inside the obstacle boundary after the planning. These results demonstrate the efficacy of the orientation planner in creating an ablated profile inside the obstacle boundary.


\section{DISCUSSION AND CONCLUSION}


In this work, we formulated an optimization model to solve the robotic laser orientation planning problem with an application of minimizing the over-cutting of healthy tissue for robotic laser surgery. We show that the laser-tissue cavity can be predicted under various incident angles and the orientation planning can be guided by the reference vector field. With this method, the surgeon can manually define a preoperative boundary or a 3D contour, and the planner can calculate an optimal incident angle to minimize the probability of cutting incorrect tissue regions during surgery.


Future works include the experiments of more obstacle vector fields and the comparisons of different gradient-based optimization solvers. Additionally, more testing for various types of phantom and ex vivo animal tissue should be conducted in anticipation of ultimate application in the clinical setting. The proposed method could also be tested with the additional variety of existing surgical lasers systems and wavelengths such as ND:YAG and Er:YAG lasers.

\bibliographystyle{IEEEtran}



\section{Appendix: Details of Derivation}

This section provides the detailed derivation for the Jacobian matrix of the angle-position kinematic model. Given a 3D surface point $q_i^k \in \R^3$ at k-th time step, this kinematic model maps an incident angle input $\theta \in \R^3$ to a new position $q^{k+1}_i \in \R^3$ and the derivation of the Jacobian matrix can be provided for fast computation. 

First, we review the kinematic model of a point robot at the surface and define a new function $Q(\cdot)$ to denote the updated point $\mathbf{q}_i^{k+1}$:  
\begin{equation*}
    \begin{aligned}
           Q(\mathbf{v(\theta)}) & = \mathbf{q}_i^{k+1} = \mathbf{q}_i^k + \mathbf{v(\theta)} * d_i
    \end{aligned}
\end{equation*}
Where $\mathbf{q}_i^{k+1}$ is the updated point from $\mathbf{q}_i^k$. The $\mathbf{q}_i^k$ is a constant in the function. The $\mathbf{v(\theta)} \in \R^3$ is a normalized incident vector with respect to the orientation angle $ \theta \in \R^3$. The $d_i$ is the penetration depth along the laser incident direction. We can write the full description of the function: 
\begin{equation*}
    \begin{aligned}
        Q(\mathbf{v(\theta})) & = \mathbf{q}_{i}^{k} + L_{G} * \mathbf{v(\theta)} * \mathrm{exp} ( \frac{s_i ^2( \mathbf{v(\theta)},  \mathbf{q}_{i}^{k} ) }{ -2 * \sigma_{G}^2} )
    \end{aligned}
\end{equation*}

Where $L_{G}$ is a fixed constant that can be learned from the data-driven model (Section II. B). The goal is to find the derivative for $Q(\mathbf{v(\theta)})$ with respect to the angle-vector variable $\mathbf{v} = \mathbf{v(\theta)}$.
\begin{equation*}
    \begin{aligned}
        \frac{\partial Q(\mathbf{v(\theta)})} {\partial \theta} & = \frac{\partial \mathbf{q_i^k}}{\partial \theta} + L_{G} * \frac{\partial }{\partial \theta} [ \mathbf{v(\theta)} * \mathrm{exp} (\frac{s_i^2( \mathbf{v(\theta), \mathbf{q_i^k} )}}{ -2 * \sigma^2_{G}} )]
    \end{aligned}
\end{equation*}

Where $\mathbf{v(\theta)}$ is the incident vector described by the angle variable $\theta$. Since $\mathbf{q_i^k}$ is a constant (the state of position from the current step), we simplify the $s_i^2( \mathbf{v(\theta), \mathbf{q}_i)}$ as $s_i^2(\mathbf{v(\theta))}$. 

We denote the incident vector as a dependent variable for the orientation angle (independent variable). Based on the chain rule, the complete derivative of the kinematic model can be described as: 
\begin{equation*}
\begin{aligned}
   \frac{\partial Q}{\partial \theta} & =  \frac{\partial Q}{\partial \mathbf{v}} * \frac{\partial \mathbf{v}}{\partial \theta} \\
   & = 
    \begin{bmatrix}
        \frac{\partial Q_1}{\partial v_1} & \frac{\partial Q_1}{\partial v_2} & \frac{\partial Q_1}{\partial v_3} \\
        \frac{\partial Q_2}{\partial v_1} & \frac{\partial Q_2}{\partial v_2} & \frac{\partial Q_2}{\partial v_3} \\
        \frac{\partial Q_3}{\partial v_1} & \frac{\partial Q_3}{\partial v_2} & \frac{\partial Q_3}{\partial v_3}  
    \end{bmatrix} *  
    \begin{bmatrix}
        \frac{\partial v_1}{\partial \theta_1} & \frac{\partial v_1}{\partial \theta_2} & \frac{\partial v_1}{\partial \theta_3} \\
        \frac{\partial v_2}{\partial \theta_1} & \frac{\partial v_2}{\partial \theta_2} & \frac{\partial v_2}{\partial \theta_3} \\
        \frac{\partial v_3}{\partial \theta_1} & \frac{\partial v_3}{\partial \theta_2} & \frac{\partial v_3}{\partial \theta_3}  
    \end{bmatrix}
\end{aligned}
\end{equation*}

Therefore, we need to determine the derivative of $\frac{\partial Q}{\partial \mathbf{v}}$ and $\frac{\partial \mathbf{v}}{\partial \theta}$ separately.

\subsection{Derivative of $\frac{\partial Q}{\partial \mathbf{v}}$}

To make the formulation more concise, we define $g_1(\mathbf{v}(\theta)) = \mathbf{v}(\theta) * \mathrm{exp}(\frac{s_i^2(\mathbf{v(\theta)})}{-2 * \sigma^2_{G}})$. We first obtain the simplified function: 
\begin{equation*}
    \begin{aligned}
      Q(\mathbf{v}) & = \mathbf{q_i^k} + L_G *  g_1(\mathbf{v}(\theta))
    \end{aligned}
\end{equation*}

\noindent
Where we first denote $\mathbf{v}(\theta) = \mathbf{v}$ for simplification. Based on the chain rule and $\mathbf{q_i^k}$ is a constant vector, we have:
\begin{equation*}
    \begin{aligned}
        \frac{Q(\mathbf{v})}{\partial \mathbf{v}} & = L_G * \frac{\partial g_1(\mathbf{v})}{\partial \mathbf{v}}
    \end{aligned}
\end{equation*}

\noindent
For $g_1(\mathbf{v})$, we have:
\begin{equation*}
    \begin{aligned}
        \frac{\partial g_1(\mathbf{v})}{\partial \mathbf{v}}   = & \frac{\partial \mathbf{v}}{\partial \mathbf{v}} * \mathrm{exp}(\frac{s_i^2(\mathbf{v})}{-2 * \sigma^2_{G}}) ~ + \\
        &  \frac{\partial s_i^2( \mathbf{v} )}{\partial \mathbf{v}}   * \mathrm{exp} (\frac{s_i^2( \mathbf{v} )}{-2 * \sigma^2_{G}})  * (\frac{1}{-2 * \sigma^2_{G}})   
    \end{aligned}
\end{equation*}

We define $g_2( \mathbf{v} ) = s^2_i(\mathbf{v})$ and the area of triangle is fixed. The $g_1( \mathbf{v})$ can be simplified as:
\begin{equation*}
\begin{aligned}
    g_1( \mathbf{v}) & = \mathbf{v} * \mathrm{exp}(\frac{g_2( \mathbf{v})}{-2 * \sigma^2_{G}})
\end{aligned}
\end{equation*}

\begin{figure}[h]
\centering
\includegraphics[scale = 0.50]{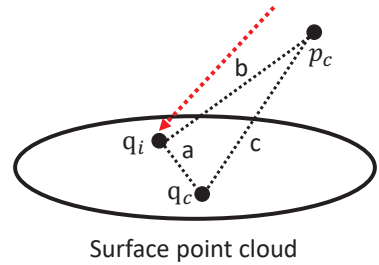}
\caption{3D triangle geometry.}
\label{figapp_1}
\end{figure}

We have $g_2(\mathbf{v}) = s_i^2(\mathbf{v})$ and $s_i(\cdot)$ is the height of the triangle contour with respect to the three sides of a, b and c, as depicted in Figure.~\ref{figapp_1}.

The area of the triangle is fixed and can be described by the three edges as $Area = \sqrt{p(p-a)(p-b)(p-c)} = 0.5 * c * s_i$. The $s_i$ is the height of the triangle. Since $g_2( \mathbf{v} )$ is a function of a, b, and c, we denote $g_2( \mathbf{v} ) = g_2 (\cdot)$ for simplification. Therefore, we can formulate an equation:
\begin{equation*}
    \begin{aligned}
        g_2( \cdot ) & = \frac{4}{c^2} * p * (p - a) * (p - b) * (p - c) 
    \end{aligned}
\end{equation*}

Where we know the edges for the triangle geometry are defined as $a(\cdot) = ||q_c - q_i||_2$, $b(\cdot) = q_i - q_c + \mathbf{v} * L_{ref}$ ($L_{ref}$ is a pre-defined constant) and $c(\cdot) = p_c - q_c = L_{ref}$. Based on the multivariate calculus, we have: 
\begin{equation*}
    \begin{aligned}
        \frac{\partial g_2( a, b, c )}{\partial v} & = \frac{\partial g_2}{\partial a} * \frac{\partial a}{\partial v} + \frac{\partial g_2}{\partial b} * \frac{\partial b}{\partial v} + \frac{\partial g_2}{\partial c} * \frac{\partial c}{\partial v} 
    \end{aligned}
\end{equation*} 

Where the $\frac{\partial g_2}{\partial a}, \frac{\partial g_2}{\partial b}, \frac{\partial g_2}{\partial c}$ can be determined by using the MATLAB symbolic math toolbox. The analytical information is used for computation (rewrite in programming) while the symbolic description is not directly applied in MATLAB. For the derivative with respect to a, b and c, we obtain:
\begin{equation*}
    \begin{aligned}
        \frac{\partial a}{\partial v} & = \frac{\partial}{\partial v} (||q_c - q_i^k||_2) = \mathbf{0 }_{3 \times 1}
    \end{aligned}
\end{equation*}
\begin{equation*}
    \begin{aligned}
        \frac{\partial b}{\partial v} & = \frac{\partial}{\partial v} (||q_i - p_c||_2)  \\ 
        & = \frac{\partial}{\partial v} (||q_i - q_c + v * L_{ref}||_2) \\ 
        & = \frac{ (q_i - q_c + v * L_{ref}) * L_{ref} }{ || q_i - q_c + v * L_{ref} ||_2 }
    \end{aligned}
\end{equation*}

\noindent
Where we know $q_c$ and $q_i^k$ are constant vectors. Since we know $c = ||p_c - q_c||_2 = L_{ref}$, we have: 
\begin{equation*}
    \begin{aligned}
        \frac{\partial c}{\partial v} & = \frac{\partial}{\partial v} (||p_c - q_c||_2) = \mathbf{0 }_{3 \times 1}
    \end{aligned}
\end{equation*}
In summary, we could derive the complete representation of the derivative information for $Q(\mathbf{v})$ with $\mathbf{Q} = \{ Q_1, Q_2, Q_3 \} \in \R^3$ and $\mathbf{\theta} = \{ \theta_1, \theta_2, \theta_3 \} \in \R^3$:
\begin{equation*}
\begin{aligned}
  \frac{\partial Q}{\partial \mathbf{v}} & = 
    \begin{bmatrix}
        \frac{\partial Q_1}{\partial v_1} & \frac{\partial Q_1}{\partial v_2} & \frac{\partial Q_1}{\partial v_3} \\
        \frac{\partial Q_2}{\partial v_1} & \frac{\partial Q_2}{\partial v_2} & \frac{\partial Q_2}{\partial v_3} \\
        \frac{\partial Q_3}{\partial v_1} & \frac{\partial Q_3}{\partial v_2} & \frac{\partial Q_3}{\partial v_3}  
    \end{bmatrix} =    
    \begin{bmatrix}
        - & \frac{\partial Q_1 }{\partial \mathbf{v}} & - \\
        - & \frac{\partial Q_2 }{\partial \mathbf{v}} & - \\
       - & \frac{\partial Q_3 }{\partial \mathbf{v}} & - \\
    \end{bmatrix}
\end{aligned}
\end{equation*}

Each component $\frac{\partial Q_i}{\partial v_j}, i, j = 1, 2, 3$ in the matrix can be derived analytically by the chain rule (shown in previous steps) and put in different objective functions. Next, we discuss the derivative of $\mathbf{v(\theta)}$ with respect to $\theta$. \\

\subsection{Derivative of $\frac{\partial \mathbf{v(\theta)}}{\partial \theta}$}

\noindent
\textbf{Rotation and Angle Model:} The incident vector is controlled the orientation angle $ \mathbf{\theta} = (\theta_1, \theta_2, \theta_3) $ in the Euclidean space. We first define a fixed initial incident vector as $\mathbf{v_0} = (v_1, v_2, v_3)$ and describe it with the rotation matrix:  
\begin{equation*}
      \begin{aligned}
      \mathbf{v(\theta)} 
      & = 
      R_x * R_y * R_z * \mathbf{v_0} \\ 
      & = 
      \begin{bmatrix}
      R_{11} & R_{12} & R_{13} \\
      R_{21} & R_{22} & R_{23} \\
      R_{31} & R_{32} & R_{33}
      \end{bmatrix} * 
      \begin{bmatrix}
      v_1 \\ 
      v_2 \\
      v_3
      \end{bmatrix} \\ 
      & = 
      \begin{bmatrix}
      R_{11} * v_1 + R_{12} * v_2 + R_{13} * v_3 \\ 
      R_{21} * v_1 + R_{22} * v_2 + R_{23} * v_3 \\
      R_{31} * v_1 + R_{32} * v_2 + R_{33} * v_3
      \end{bmatrix} \\
      & = 
      \begin{bmatrix}
       v_1^* \\
       v_2^* \\
       v_3^* 
      \end{bmatrix}
\end{aligned}
\end{equation*}

The $\mathbf{v_0}$ can be defined as an initial incident vector and this vector will be updated by the changes of orientation angles. For this description, we have $R_x$:
\begin{equation*}
    \begin{aligned}
      R_x = 
      \begin{bmatrix}
      1 & 0 & 0 \\
      0 & cos(\theta) & -sin(\theta) \\
      0 & sin(\theta) & cos(\theta) 
      \end{bmatrix}
    \end{aligned}
\end{equation*}

\noindent
and we have $R_y$:
\begin{equation*}
    \begin{aligned}
     R_y =  \begin{bmatrix} 
      cos(\theta) & 0 & sin(\theta) \\
      0 & 1 & 0 \\
      -sin(\theta) & 0 & cos(\theta)
      \end{bmatrix}
    \end{aligned}
\end{equation*}

\noindent
and finally we obtain $R_z$: 
\begin{equation*}
    \begin{aligned}
    R_z = \begin{bmatrix} 
      cos(\theta) & -sin(\theta) & 0 \\
      sin(\theta) & cos(\theta) & 0 \\
      0 & 0 & 1
      \end{bmatrix}
    \end{aligned}
\end{equation*}
      
\noindent
The derivative with respect to the $\mathbf{\theta}$ can be described as: 
\begin{equation*}
    \begin{aligned}
      \frac{\partial \mathbf{v} (\theta)}{\partial \theta} & = 
      \begin{bmatrix}
          \frac{\partial v_1^*}{\partial \theta_x} &  \frac{\partial v_1^*}{\partial \theta_y} &  \frac{\partial v_1^*}{\partial \theta_z} \\ 
          \frac{\partial v_2^*}{\partial \theta_x} &  \frac{\partial v_2^*}{\partial \theta_y} &  \frac{\partial v_2^*}{\partial \theta_z} & \\
          \frac{\partial v_3^*}{\partial \theta_x} &  \frac{\partial v_3^*}{\partial \theta_y} &  \frac{\partial v_3^*}{\partial \theta_z} 
      \end{bmatrix} = 
      \begin{bmatrix}
      | &  | & | \\ 
      \frac{\partial \mathbf{v}}{\partial \theta_x } &  \frac{\partial \mathbf{v}}{\partial \theta_y } & 
      \frac{\partial \mathbf{v}}{\partial \theta_z }  \\ 
      | & | & |  
      \end{bmatrix}
    \end{aligned}
\end{equation*}

\noindent
For each column, we obtain:

\begin{equation*}
    \begin{aligned}
      \frac{\partial \mathbf{v}}{\partial \theta_j} & = \frac{\partial (R_{xyz})}{\partial \theta_j} * \mathbf{v_0} 
    \end{aligned}
\end{equation*}

Where we have $R_{xyz} = R_x * R_y * R_z$ and $\frac{\partial (R_{xyz})}{\partial \theta_j}$ is an item-wise derivative. For example, the derivative of the i-th item in $\mathbf{v}$ and j-th $\theta_j$ element is denoted as:
\begin{equation*}
\begin{aligned}
      \frac{\partial ( v_i^* ) }{\partial \theta_j} 
     & = \frac{\partial R_{i,1}}{\partial \theta_j} * v_1 + \frac{\partial R_{i,2}}{\partial \theta_j} * v_2 + \frac{\partial R_{i,3}}{\partial \theta_j} * v_3 
     \\ 
     & = 
     \begin{bmatrix}
     \frac{\partial R_{i,1}}{\partial \theta_j} & \frac{\partial R_{i,2}}{\partial \theta_j} & \frac{\partial R_{i,3}}{\partial \theta_j} 
     \end{bmatrix} * 
     \begin{bmatrix}
     v_1 & v_2 & v_3
     \end{bmatrix}^T
\end{aligned}
\end{equation*}

Therefore, we can find the derivative of the multiplied rotation matrix by taking the derivative of each item to the $(\theta_j,~j = x, y, z)$ in the 3D space. 

In summary, we review the complete derivative information of the kinematic system between the 3-DOF laser incident angle (input) and the 3-DOF predicted position (output). We use the MATLAB symbolic math toolbox to take the derivatives of some of the complex functions. The derivative of the kinematic model can be derived analytically and thus we can achieve the fast computation in different software platform. 


\clearpage
\newpage

\end{document}